%% file: 0_main.tex
\documentclass[11pt,a4paper]{article}
\usepackage[hyperref]{acl2020}
\usepackage{times}
\usepackage{latexsym}

\usepackage{url}

\usepackage{hyperref}
\usepackage{dirtytalk}
\usepackage{url}
\usepackage{graphicx} 
\usepackage{textcomp}
\usepackage{float}
\usepackage{subcaption}

\aclfinalcopy % Uncomment this line for the final submission

\title{Shape of synth to come: Why we should use synthetic data for English surface realization}

\author{Henry Elder \\
  ADAPT Centre \\
  Dublin City University \\
  {\tt henry.elder@adaptcentre.ie  } \\\And
  Robert Burke \\
  \texttt{sharpobject@gmail.com} \\\AND
  Alexander O'Connor \\
  Autodesk, inc. \\
  {\tt alex.oconnor@autodesk.com} \\\And
  Jennifer Foster  \\
  School of Computing \\
  Dublin City University \\
  {\tt  jennifer.foster@dcu.ie} \\}

\date{}

\begin{document}
\maketitle

\input{tex_files/abstract.tex}
\input{tex_files/introduction_jf.tex}
\input{tex_files/system.tex}
\input{tex_files/results.tex}

\input{tex_files/conclusion.tex}
\section*{Acknowledgments}
We thank the anonymous reviewers for their helpful comments. This research is supported by Science Foundation Ireland in the ADAPT Centre for Digital Content Technology. The ADAPT Centre for Digital Content Technology is funded under the SFI Research Centres Programme (Grant 13/RC/2106) and is co-funded under the European Regional Development Fund.

\bibliography{references}
\bibliographystyle{acl_natbib}

\end{document}

%% file: tex_files/abstract.tex
\begin{abstract}
    %The 2019 Surface Realization shared task introduced a new rule prohibiting the use of synthetically created data. 
    %  
    %This rule was based on incomplete information from the 2018 shared task.
    % and therefore the results of the 2019 shared task may be misleading.
    %
    %
    % This paper demonstrates how, with synthetic data, 
    %Contrary to prior work, we find that, with synthetic data, performance on the 2018 shared task dataset improves from 72.7 BLEU to 80.1.
    % We demonstrate that with synthetic data the state-of-the-art is improved from 72.7 BLEU to 80.1.
    %
    %It is our opinion, supported by experiments, that banning the use of synthetic data in the shared task was potentially unnecessary and could be to the detriment of future research.
    % We conduct analysis on the effects synthetic data has on performance, and thus contend that incorporating synthetic data into training is key to achieving the quality necessary for production ready systems.
    %%%%%EDITED BY JENNIFER
    The Surface Realization Shared Tasks of 2018 and 2019 were Natural Language Generation shared tasks with the goal of exploring approaches to surface realization from Universal-Dependency-like trees to surface strings for several languages. In the 2018 shared task there was very little difference in the absolute performance of systems trained with and without additional,  synthetically created data, and a new rule prohibiting the use of synthetic  data was introduced for the 2019 shared task. Contrary to the findings of the 2018 shared task, we show, in experiments on the English 2018 dataset, that the use of synthetic data can have a substantial positive effect -- an improvement of almost 8 BLEU points for a previously state-of-the-art system.  We analyse the effects of synthetic data, and we argue that its use should be encouraged rather than prohibited so that future research efforts continue to explore  systems that can take advantage of such data.
\end{abstract}

% The Surface Realization Shared Task is a Natural Language Generation task with rising popularity, increasing from 9 teams in 2018 to 14 teams in 2019. However, the 2019 shared task introduced a new rule prohibiting the use of synthetic data. We demonstrate that with synthetic data the state-of-the-art is improved from 72.7 BLEU to 80.1. We conduct analysis on the effects synthetic data has on performance, and thus contend that incorporating synthetic data into training is key to achieving the quality necessary for production ready systems.

%% file: tex_files/introduction_jf.tex
\section{Introduction}
\label{sec:introduction}

\begin{figure}
    \centering
    \includegraphics[width=0.6\linewidth]{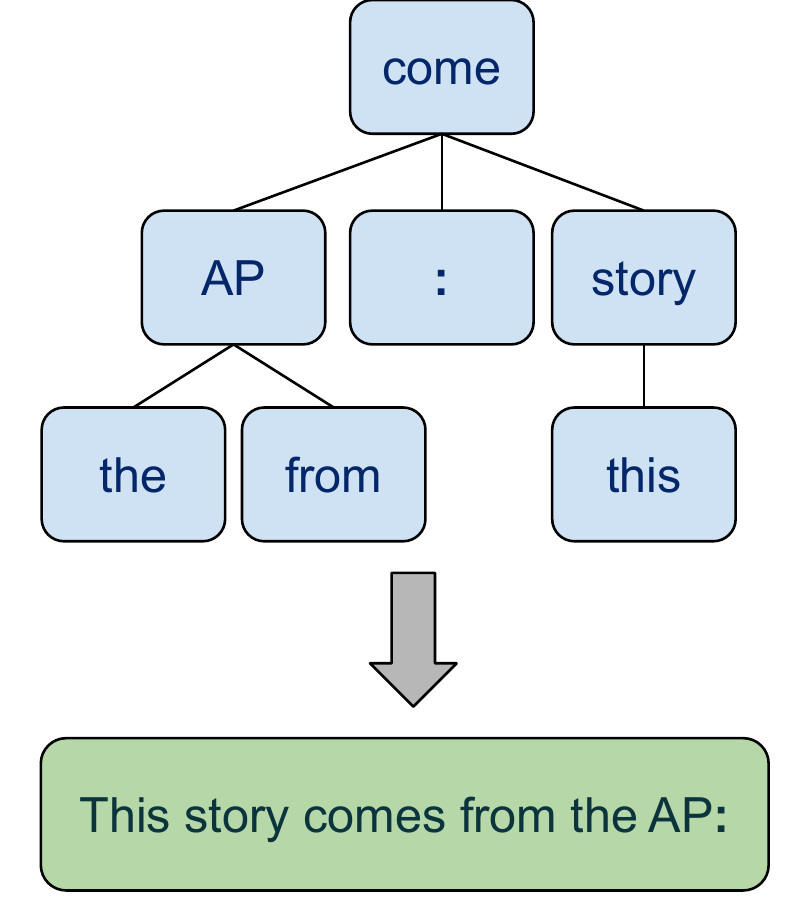}
    \caption{Example tree and reference sentence}
    \label{fig:surface_realization_example}
\end{figure}

% Topic - The task itself
The shallow task of the recent surface realization (SR) shared tasks \citep{first_surface_realization_shared_task, mulit_surface_realization_2018, second_surface_realization_2019} appears to be a relatively straightforward problem.  
Given a tree of lemmas, a system has to restore the original word order of the sentence and inflect its lemmas, see Figure \ref{fig:surface_realization_example}. 
% Question - How to get the best performance 
Yet SR systems often struggle, even for a relatively fixed word order language such as English.
% Significance - More complex forms of the task that might be more useful in practice
Improved performance would facilitate investigation of more complex versions of the shallow task, such as the deep task in which function words are pruned from the tree, which may be of more practical use in pipeline natural language generation (NLG) systems \citep{moryossef-etal-2019-step, elder-etal-2019-designing, castro-ferreira-etal-2019-neural}. 
% Improved performance would facilitate investigation of more complex versions of this problem,

% Preamble / claim(?) - Use of synthetic data
In this paper we explore the use of synthetic data for the English shallow task.
Synthetic data is created by taking an unlabelled sentence, parsing it with
an open source universal dependency parser\footnote{A number of these exist, e.g.  \url{https://github.com/stanfordnlp/stanfordnlp} and  \url{http://lindat.mff.cuni.cz/services/udpipe/}} and transforming the result into the input representation.

% Claim - Extra data is really important for this task and the quality
Unlike in the 2018 shared task, where a system trained with synthetic data performed roughly the same as a system trained on the original dataset \citep{elder_srst_2018, king_white_srst_2018}, we find its use leads to a large improvement in performance.
% Evidence - Scores are much higher
The state-of-the-art on the dataset is 72.7 BLEU-4 score \citep{yu-etal-2019-head} -- our system achieves a similar result of 72.3, which improves to 80.1 with the use of synthetic data.
% Reason - Analyse what the extra data helps to improve
We analyse the ways in which synthetic data helps to improve performance, finding that longer sentences are particularly improved and more exactly correct linearizations are generated overall.
% Probably not necessary but could mention the other fields like text classification / sequence labelling (unsupervised language model pretraining) and machine translation (synthetic data using back translation)

% Preamble / Claim - Not allowed to use additional data this year
% Then we talk about the question of additional data
Although it is common knowledge that machine learning systems typically benefit from more data, this 7.4 point jump in BLEU is important and worth emphasizing. The 2019 shared task introduced a new rule which prohibited the use of synthetic data. This was done in order to make the results of different systems more comparable.
% s that by having all systems train on the same set of data results would be more comparable. 
%
However, systems designed with smaller datasets in mind might not scale to the use of synthetic data, and an inadvertent consequence of such a rule is that it may produce results which could be misleading for future research directions.

For instance, the system which was the clear winner of  this year's shared task \cite{ims_srst_2019} used tree-structured long short-term memory (LSTM) networks \citep{tai_tree_lstm_2015}.
In general, tree LSTMs can be slow and difficult to train.\footnote{\url{https://github.com/dasguptar/treelstm.pytorch/issues/6}}
\citet{graphAMR_ACL2018} utilized a variant of the tree LSTM in a similar NLG task, converting abstract meaning representation (AMR) graphs to text.
Following the state-of-the-art system \citep{neural_amr_konstas_ACL_2017}, which used standard LSTMs, \citeauthor{graphAMR_ACL2018} augmented their training with synthetic data.
Though their system outperformed \citeauthor{neural_amr_konstas_ACL_2017} at equivalent levels of additional training sentences, it was unable to scale up to the 20 million sentences used by the best \citeauthor{neural_amr_konstas_ACL_2017} system and ultimately did not outperform them.\footnote{\citeauthor{graphAMR_ACL2018}'s best system achieved 33.0 BLEU score with 2 million additional sentences, while \citeauthor{neural_amr_konstas_ACL_2017} scored 32.3 with 2 million and 33.8 with 20 million (the best overall system).}

% More significance - Ehud Reiter and the quality
Critics of neural NLG approaches\footnote{See, for example, \url{https://ehudreiter.com/}} emphasise that quality and reliability are at the core of production-ready NLG systems.
What we are essentially arguing is that if using synthetic data contributes to producing higher quality outputs, then we ought to ensure we are designing systems that can take advantage of synthetic data.

% Say something about the different languages. That although we test only on english the system is designed to take input in the form of UD trees, which are universal. 

%% file: tex_files/system.tex
\section{System Description}
\label{sec:system}

\subsection{Data}

We evaluate on the Surface Realization Shared Task (SRST) 2018 dataset \citep{mulit_surface_realization_2018} for English\footnote{\url{http://taln.upf.edu/pages/msr2018-ws/SRST.html}}, which was derived from the Universal Dependency English Web Treebank 2.0\footnote{\url{https://github.com/UniversalDependencies/UD_English-EWT}}. The training set consists of 12,375 sentences, dev 1,978, test 2,062.
% More description on how it was derived? Kind of already covered elsewhere and in the introduction

% Introduce baseline system
\subsection{Baseline system}
The system we use is an improved version of a previous shared task participant's system \cite{elder_srst_2018}.
This baseline system is a bidirectional LSTM encoder-decoder model.
The model is trained with copy attention \citep{Vinyals2015,See2017} which allows it to copy unknown tokens from the input sequence to the output. 
The system performs both linearization and inflection in a single decoding step.
To aid inflection, a list is appended to the input sequence containing possible forms for each relevant lemma.

Depth first linearization \citep{neural_amr_konstas_ACL_2017} is used to convert the tree structure into a linear format, which is required for the encoder.
This linearization begins at the root node and adds each subsequent child to the sequence, before returning to the highest node not yet added.
Where there are multiple child nodes one is selected at random.
Decoding is done using beam search, the output sequence length is artificially constrained to contain the same number of tokens as the input.

\subsection{Improvements to baseline}\label{baseline:improvements}

\paragraph{Random linearizations}
In the baseline system, a single random depth first linearization of the training data is obtained and used repeatedly to train the model.
Instead, we obtain multiple linearizations, so that each epoch of training data potentially contains a different linearization of the same dependency tree.
This makes the model more robust to different linearizations, which is helpful as neural networks don't generally deal well with randomness \citep{slug2slug_naacl_2018}.

% scoping brackets?
\paragraph{Scoping brackets}
Similar to \citet{neural_amr_konstas_ACL_2017} we apply scoping brackets around child nodes. This provides further indication of the tree structure to the model, despite using a linear sequence as input.

\paragraph{Restricted beam search}
In an attempt to reduce unnecessary errors during decoding, our beam search looks at the input sequence and restricts the available vocabulary to only tokens from the input, and tokens which have not yet appeared in the output sequence.
This is similar to the approach used by \citet{king_white_srst_2018}.
% %
% We go into further detail on the restricted beam search in Appendix \ref{sec:restricted_beam_search}.

\subsection{Synthetic Data}

To augment the existing training data we create synthetic data by parsing sentences from publicly available corpora.
The two corpora we investigated are Wikitext 103 \citep{DBLP:conf/iclr/MerityX0S17} and the CNN stories portion of the DeepMind Q\&A dataset \citep{NIPS2015_5945}.

% The processing
Each corpus requires some cleaning and formatting, after which they can be sentence tokenized using CoreNLP \cite{manning-EtAl:2014:P14-5}.
Sentences are filtered by length -- min 5 tokens and max 50 -- and for vocabulary overlap with the original training data -- set to 80\% of tokens in a sentence required to appear in the original vocabulary.
These sentences are then parsed using the Stanford NLP UD parser \citep{qi2018universal}.
This leaves us with 2.4 million parsed sentences from the CNN stories corpus and 2.1 million from Wikitext.

It is a straightforward process to convert a parse tree into synthetic data.
First, word order information is removed by shuffling the IDs of the parse tree, then the tokens are lemmatised by removing the form column.
This is the same process used by the shared task organizers to create datasets from the UD treebanks.

While it has been noted that the use of synthetic data is problematic in NLG tasks (WeatherGov \citep{Liang2009LearningSupervision} being the notable example) our data is created differently.
The WeatherGov dataset is constructed by pairing a table with the output of a rule-based NLG system. 
This means any system trained on WeatherGov only re-learns the rules used to generate the text.
Our approach is the reverse; we parse an existing, naturally occurring sentence, and, thus, the model must learn to reverse the parsing algorithm.
%
% This turns out to be a useful task than training on data from the rule-based system.

\subsection{Training}
The system is trained using a custom fork\footnote{\url{https://github.com/Henry-E/OpenNMT-py}} of the OpenNMT-py framework \citep{opennmt_py_2017}, the only change made was to the beam search decoding code. Hyperparameter details and replication instructions are provided in our project's repository\footnote{\url{https://github.com/Henry-E/surface-realization-shallow-task}}, in particular in the config directory. 

Vocabulary size varies based on the datasets in use. It is determined by using any tokens which appears 10 times or more.
When using the original shared task dataset, the vocabulary size is 2,193 tokens, training is done for 33 epochs and takes 40 minutes on two Nvidia 1080 Ti GPUs.
All hyperparameters stay the same when training with the synthetic data, except for vocabulary size and training time. For the combined shared task, Wikitext and CNN datasets the vocabulary size is 89,233, training time increases to around 2 days, and uses 60 random linearizations of the shared task dataset and 8 of the Wikitext and CNN datasets.

\subsection{Evaluation}
The evaluation is performed on detokenized sentences\footnote{Using detokenized inputs for BLEU makes the score very sensitive to detokenization used and in the 2019 shared task evaluation was changed to use tokenized inputs instead.} using the official evaluation script from the 2018 shared task. We focus on BLEU-4 score \citep{Papineni:2002:BMA:1073083.1073135} which was shown in both shared tasks to be highly correlated with human evaluation scores.

% In this section we describe the changes we made to one of the best performing system for the 2018 shared task \cite{Elder2018GeneratingModels}.

% The base model is an bi-directional LSTM with copy attention. Lexical features such as part-of-speech and dependency relation are given their own feature vectors and concatented at input with the word vectors for lemmas.

% The model learns to automatically perform inflection, to aid in this we also append a list of suggested forms for lemmas to the end of the sequence.

% randomised inputs
% To make the system (more robust to different linearizations), the input trees were randomly linearized. In the previous system this was done once and the system trained on repeated batches on the same random linearization. In this work we randomly linearize the dataset at every epoch / repetition of data.

% restricted decoding
% We included an additional feature from another top performing system (Mike White's paper) which restricts output sequences to include only tokens that appeared in the input. 
% TODO add an appendix describing the bipartite matching code.

% The additional data used

%% file: tex_files/results.tex
\section{Results}
\label{sec:results}

% \subsection{Results}

In Table \ref{tab:test_set_only_srst_training}, we compare our results on the test set with those reported in \citet{yu-etal-2019-head}, which include the \citeauthor{yu-etal-2019-head} system (Yu19), the best 2018 shared task result for English \citep{elder_srst_2018} (ST18) and \citeauthor{yu-etal-2019-head}'s implementation of two other baselines, \citet{bohnet-etal-2010-broad} (B10)  and \citet{puduppully_linearization_2016} (P16) .
Ignoring for now the result with synthetic data, we can see that our system is competitive with that of Yu et al (72.3 vs 72.7).
% \usepackage{graphicx}
\begin{table}[]
\begin{tabular}{ll}
\hline
                                                            & BLEU-4   \\ \hline
B10                                                         & 70.8    \\
P16                                                         & 65.9   \\
ST18                                                        & 69.1    \\
Yu19                                                        & \textbf{72.7}    \\
Ours                                                        & 72.3 \\  \hline
Ours + Synthetic data                                            & \textbf{80.1} \\ \hline

\end{tabular}
\caption{Test set results for baselines trained on the original dataset and the final model which uses synthetic data}
\label{tab:test_set_only_srst_training}
\end{table}

%\subsection{Ablation analysis}
In Section~\ref{baseline:improvements}, we described three improvements to our baseline system: random linearization, scoping and restricted beam search. An ablation analysis of these improvements on the dev set is shown in Table \ref{tab:ablation_baseline}. The biggest improvement comes from the introduction of random linearizations. However, all three make a meaningful, positive contribution.
%We present an ablation analysis of usage of the synthetic data in Table \ref{tab:dataset_ablation_analysis}.
%

% \usepackage{graphicx}
\begin{table}[]
% \centering
\resizebox{\linewidth}{!}{%
\begin{tabular}{ll}
\hline
System                                     & BLEU-4 \\ \hline
SR Baseline                                & 57.3  \\
SR + Random Lins                           & 65.1 \\
SR + Random Lins + Scope                   & 69.2    \\
SR + Random Lins + Scope + Restricted Beam & 72.2    \\ \hline
\end{tabular}%
}
\caption{Dev set results for ablation of the baseline system plus improvements, trained only on the original dataset}
\label{tab:ablation_baseline}
\end{table}

% \usepackage{graphicx}
\begin{table}[t]
% \centering
\resizebox{0.7\linewidth}{!}{%
\begin{tabular}{ll}
\hline
Data used            & BLEU-4 \\ \hline
Improved SR Baseline (SRST) & 72.2  \\
SR + Wikitext            & 79.8   \\
SR + CNN             & 80.3   \\
SR + CNN + Wikitext      & 80.8    \\ \hline
\end{tabular}%
}
\caption{Dev set results for the SR shared task data with additional synthetic data: the role of the corpus}
\label{tab:dataset_ablation_analysis}
\end{table}

\subsection{The Effect of Synthetic Data}
The last row of Table~\ref{tab:test_set_only_srst_training} shows the effect of adding synthetic data. BLEU score on the test set jumps from 72.3 to 80.1.
To help understand why additional data makes such a substantial difference, we perform various analyses on the dev set, including examining the effect of the choice of unlabeled corpus and highlighting interesting differences between the systems trained with and without the synthetic data.

\paragraph{The role of corpus}
Table~\ref{tab:dataset_ablation_analysis} compares the Wikitext corpus as a source of additional training data to the CNN corpus. Both the individual results and the result obtained by combining the two corpora show that there is little difference between the two.

\begin{figure}
    \centering
    \includegraphics[width=0.8\linewidth]{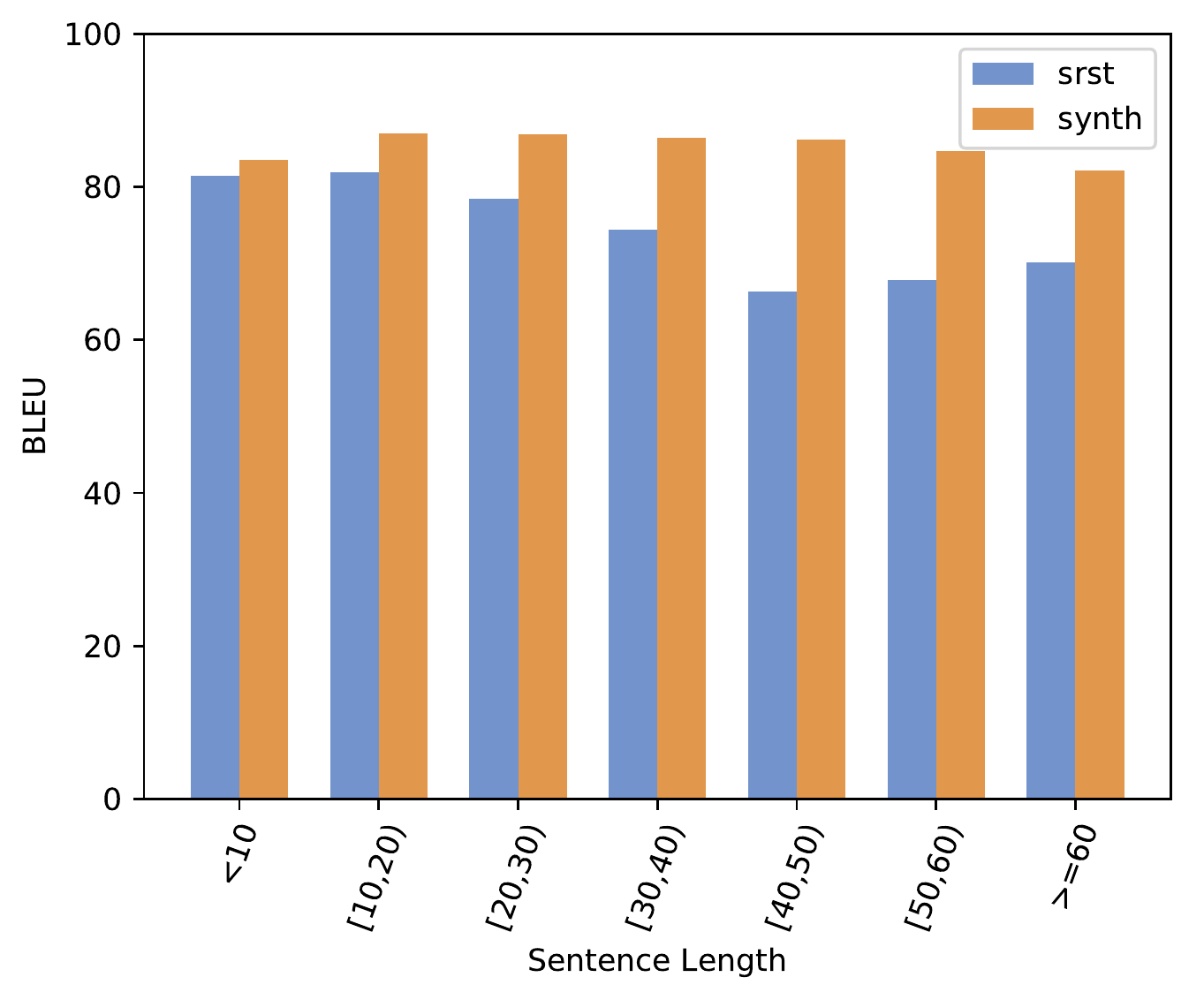}
    \caption{BLEU score breakdown by sentence length buckets, comparing our best model trained on the original dataset with one trained with synthetic data}
    \label{fig:bleu_sent_buckets}
\end{figure}

\paragraph{Sentence length and BLEU score}
Using compare-mt \citep{neubig19naacl} we noticed a striking difference between the systems with regards to performance on sentences of different length.\footnote{These are results for the tokenized versions of the generated and reference sentences, hence the higher numbers.} This is shown in Figure \ref{fig:bleu_sent_buckets}.
%

%

%
%This was particularly surprising 
Even though the synthetic data sentences were limited to 50 tokens in length, the synthetic data  performed equally well for sentence length buckets 50-60 and 60+, while the baseline data system performed relatively worse. It is  possible this is due to the synthetic data system containing a larger vocabulary and being exposed to a wider range of commonly occurring phrases, which make up parts of longer sentences.
% Further work is needed to understand the possible mechanisms that lead to this effect.

% \paragraph{Dependency relations and bleu score}
% We investigate the correlation between the number of times a dependency relation appeared in a sentence and bleu score of that sentence. And then compared the results between the two systems. 

% \paragraph{Only modelling linearization}
% Another hypothesis we had was that much of the performance improvement was coming from improved linearization, rather than inflection, so we ran an experiment to test the relative performance of the linearization.

% We present results both with and without additional data.

% (tree lstm INLG paper) presents state-of-the-art results on English for the SRST-18 test set and compares with a number of others approaches. We match their results when training on the same dataset and far exceed it when training with additional data. 

% While it is unsurprising that training with additional data increases performance, we demonstrate that the difference in performance is a least partly linked to the decrease in performance over longer sequences demonstrated in figure x.

\begin{table}[]
\resizebox{0.8\linewidth}{!}{%
\begin{tabular}{lll}
\hline
                                & SRST & Synth \\ \hline
Exact match                     & 1159 & 1314  \\
+ Punctuation error only                & 43   & 46    \\
+ Inflection error only                 & 123  & 142   \\ \cline{2-3}
Total (relatively error free)                           & 1325 & 1502 \\
Remaining errors                       & 653  & 476   \\ \hline
\end{tabular}%
}
\caption{Error analysis breakdown for the 1,978 dev sentences. SRST is our system without synthetic data and Synth is our system with synthetic data.}
\label{tab:error_analysis}
\end{table}

\paragraph{Error Analysis}
%\label{sec:manual_error_analysis}
We perform some preliminary analysis that could serve as a precursor to more detailed human evaluation.
% simplify / reduce the amount of human evaluation necessary.
%
Table \ref{tab:error_analysis} lists the number of exact matches, in which the tokenized reference sentence and the generated sentence exactly match.
We also detect relatively minor errors, namely punctuation and inflection, in which these are the only differences between the reference and generated sentences.
Punctuation errors are typically minor and there is usually ambiguity about their placement.\footnote{In the 2019 shared task an additional feature was provided to indicate the position of punctuation relative to its head token.}
Inflection errors occur when a different inflected form has been chosen by the model than in the reference sentence.
These tend to be small differences and are often valid alternatives, e.g. choosing \textit{'m} over \textit{am}.
%
% In this section we look at breakdown of the performance of the different models in terms of
% \begin{enumerate}
%     \item Identically expressed sentences
%     \item Sentences containing only inflection errors
%     \item Sentences containing only punctuation errors
%     \item Linearization errors
%     \begin{enumerate}
%         \item Still Plausable linearizations
%         \item Unnatural / unreadable linearizations
%     \end{enumerate}
% \end{enumerate}

Within the remaining uncategorized sentences are mostly linearization errors.
Linearization errors come in two main categories; non-breaking, in which the linearization is different from the reference sentence but is still valid and communicates the same meaning as the reference -- see Example 1 below; and breaking, where the linearization has clear errors and doesn't contain the same meaning as the reference sentence -- see Example 2 below.

\begin{enumerate}
    \item Non-breaking
    \begin{enumerate}
        \item Ref: From the AP comes this story:
        \item Synth: This story comes from the AP:
    \end{enumerate}
    \item Breaking
    \begin{enumerate}
        \item Ref: I ran across this item on the Internet.
        \item Synth: I ran on the internet across this item.
    \end{enumerate}
\end{enumerate}
%
% In future work, we aim to obtain funding for crowd sourcing human evaluations of these kinds of sentences.

This kind of breakdown in an error analysis may help understand the quality of these systems in more absolute terms, since it's the overall number of accurate sentences which matters.
This could be more intuitive than comparing BLEU scores relative to prior models when deciding whether to apply a system in a business setting.

% Using an error analysis with absolute numbers may help understand the quality of these systems, more so than using relative performance measures. Since in a business setting it matters what the overall number of accurate sentences are, not how were in 

%% file: tex_files/conclusion.tex
\section{Conclusion}
\label{sec:conclusion}
We have argued for the use of synthetic data in English surface realization, justified by the fact that its use 
 gives a significant performance boost on the shallow task, from 72.7 BLEU up to 80.1. While this is not yet at the level of reliability needed for neural NLG systems to be used commercially, it is a step in the right direction. 
 
 Assuming the use of synthetic data, more needs to be investigated in order to fully maximize its benefit on performance. Future work will look more closely at the choice of corpus, construction details of the synthetic dataset, as well as the trade-off between training time and accuracy that comes with larger vocabularies.
 
 The work described in this paper has focused on English. Another avenue of research would be to investigate the role of synthetic data in surface realization in other languages.